\documentclass[10pt]{article}

\usepackage[superscript,sort]{cite}

\usepackage{ifthen,ifpdf}
\ifpdf
	\usepackage[pdftex]{hyperref}	 \hypersetup{colorlinks=true,linkcolor=black,citecolor=black,urlcolor=blue,backref=page,bookmarks=true,breaklinks=true,plainpages=false }
	\usepackage[pdftex]{graphicx}
	\usepackage[usenames,pdftex]{color}
\else
	 \usepackage[colorlinks=true,breaklinks=true,plainpages=false]{hyperref}
	\usepackage{graphicx}
	\usepackage[usenames]{color}
\fi

\usepackage{amsthm,amscd,amsxtra,amsfonts,amsmath,amssymb,multirow}
\usepackage{wrapfig}
\usepackage[footnotesize]{caption}
\usepackage[tiny,compact]{titlesec}
\usepackage[textwidth=0.8in,textsize=footnotesize]{todonotes}
\usepackage{comment}
\usepackage{color}
\usepackage{subcaption}

\usepackage{booktabs}
\usepackage{threeparttable}

\usepackage{algorithm}
\usepackage[noend]{algpseudocode}

\usepackage{helvet}

\usepackage[letterpaper,top=0.5in,bottom=0.5in,left=0.5in,right=0.5in]{geometry}

\titlespacing*{\section}{0pt}{*0}{*0}
\titlespacing*{\subsection}{0pt}{*0}{*0}
\titlespacing*{\subsubsection}{0pt}{*0}{*0}
\titlespacing{\paragraph}{0pt}{*0}{*1}
\newcommand{\T}{\mathcal{T}}

\newcommand{\R}{\mathbb{R}}

\newcommand{\proj}{\mathbf{proj}}
\newcommand{\sgn}{\mathrm{sign}}

\newcommand{\W}{\tilde{W}}

\newtheorem{theorem}{Theorem}[section]

\begin{document}


\title{Deep Learning for Real-Time Crime Forecasting and Its Ternarization}
\author{Bao Wang${}^\dag$\thanks{corresponding author}, Penghang Yin${}^\dag$, \\
Andrea L. Bertozzi${}^\dag$, P. Jeffrey Brantingham${}^\ddag$, Stanley J. Osher${}^\dag$ and Jack Xin${}^\S$
\footnote{
Address correspondences  to Bao Wang. E-mail: wangbaonj@gmail.com}  \\
$\dag$ Department of Mathematics, UCLA\\
$\ddag$ Department of Anthropology, UCLA\\
$\S$ Department of Mathematics, UCI\\
}

\date{\today}

\maketitle
\begin{abstract}
Real-time crime forecasting is important. However, accurate prediction of when and where the next crime will happen is difficult.  No known physical model provides a reasonable approximation to such a complex system. Historical crime data are sparse in both space and time and the signal of interests is weak. In this work, we first present a proper representation of crime data. We then adapt the spatial temporal residual network on the well represented data to predict the distribution of crime in Los Angeles at the scale of hours in neighborhood-sized parcels. These experiments as well as comparisons with several existing approaches to prediction demonstrate the superiority of the proposed model in terms of accuracy. Finally, we present a ternarization technique to address the resource consumption issue for its deployment in real world. This work is an extension of our short conference proceeding paper [Wang et al, Arxiv 1707.03340].
\end{abstract}


\vskip 0.5cm
\section{Introduction} \label{Introducation}
Forecasting crime at hourly or even finer temporal scales in micro-geographic regions is an important scientific and practical problem. Anticipating where and when crime is most likely to occur creates novel opportunities to prevent crime. However, accurate crime forecasting at fine spatial temporal scales is very challenging. The occurrence of  crime depends on complex factors, many of which cannot be described quantitatively. Statistically, crime is extremely stochastic and sparse in both space and time \cite{RN3516}. Recent efforts have been devoted to the mathematical and statistical modeling of crime. Short et al introduced a novel partial differential equations (PDE) model to simulate crime hotspots and analyzed the regime for different dynamical patterns \cite{Short:2008M3AS, Short:2010PNAS, Short:2010SIAMADS}. The PDE model provides a macroscale description which can be regarded as a continuum limit of the microscopic random walk.
Considering crime as self-exciting, Mohler et al adapted the epidemic type aftershock sequence (ETAS) model to crime modeling \cite{Mohler:2011JASA, Short:2014DCDSB}. The ETAS model provides a microscopic representation of the crime events with predictive power. Such point process idea have been extended to  study other crime problems such as crime missing data reconstruction \cite{Stomakhin:2011InverseProblem}. Another class of crime predictors uses autoregressive integrated moving average (ARIMA) or other simple statistical models \cite{Chen:2008ARIMA, Gerber:2014}. The aforementioned models are built only on historical data. There is also interesting work on crime prediction using social network data, e.g., Twitter \cite{XiaofengWang:2012ICSC, Chen:2015Crime}.

Deep learning has recently been used for crime modeling and forecasting. In our previous work, we considered real-time crime forecasting at fine spatial scale \cite{Wang:2017Crime1}. Kang et al studied the crime forecasting problem by transforming it into binary classification problem \cite{Kang:2017Crime}. The key idea is to have a convolutional neural network (CNN) learn the features for crime forecasting with inputs of historical data, weather, geographical information, etc. Finally, they apply a support vector machine (SVM) to classify the region into crime or no crime with a posterior probability. This is an interesting idea, but not an optimal approach. Consider two regions. One region always has one crime happen with certainty. The other region has many crimes to happen, but only with 90 percent probability. Based on the classification approach, the first region would be flagged for patrol over the second. This model does not fully model the fine scale spatial temporal patterns in the crime data.

Recent advances in deep learning techniques has made forecasting of complex spatial temporal crime patterns more tractable \cite{LeCun:2015Nature, Hochreiter:1997NC, Junbo:2017, Jain:2016, Holden:2017, Li:2016GatedGraphNN}. Some of the most successful applications include citywide traffic flow forecasting, motion prediction  and human object interaction modeling. Zhang et al \cite{Junbo:2017} create an ensemble of residual networks \cite{He:2016ResNet} to study the traffic flow, their network is called ST-ResNet.  The key idea is to map the traffic flow at each time slot to an image and explicit specify the dependencies. Their model gave excellent traffic flow forecasting in Beijing and New York city. Jain et al \cite{Jain:2016} proposed a jointly trainable neural network structure, called a structural recurrent neural network (SRNN), which is a feed-forward arrangement of RNN units. The SRNN gives state-of-the-art motion forecasting. Moreover, the SRNN is scalable to massive data sets. For periodic motion forecasting, Holden et al \cite{Holden:2017} proposed a phased-function neural network for character control, their techniques have been successfully used in the gaming industry.

Despite CNNs' superior performance in various real-world applications including crime prediction, their memory and energy consumptions can be a problem, especially when deployed on mobile devices with limited resources, due to the huge number of floating-point parameters in the models. Recent efforts have been made to develop quantization techniques \cite{bc_15,xnornet_16,bnn_16,twn_16,ttq_16,twn_17} for training CNNs with low precision parameters. Thus we are able to compress the model size and speed up computation during inference. For example, in binary weight neural networks (BNNs) \cite{bc_15,xnornet_16,bnn_16}, the weights in the same fully-connected or convolutional layer are restricted to have the same magnitude. For a layer with $n$ binary weights, the storage of these parameters only requires the memory for one 32-bit floating-point number and $n$ 1-bit binary numbers (i.e., $\pm 1$) instead of that for $n$ 32-bit weights, resulting in approximately $32\times$ memory savings. Moreover, at inference time, the need for floating-point multiplications can be eliminated by leveraging the distributive law during forward propagation, which enables faster deployment and substantial energy savings. More precisely, in BNNs, a weight filter matrix or a 4 dimensional tensor, can be expressed as

$$W=\alpha B,$$

where $\alpha>0$ is the layer-wise scaling factor and $B$ has the same size as $W$ but only contains entries $\pm 1$. Given input $I$, the forward propagation calls for evaluating

$$W\ast I = \alpha (B \ast I), $$

where $\ast$ denotes the convolution operation or matrix-vector multiplication. Note that the computation of $B \ast I$ involves additions and subtractions only. Unfortunately, weight binarization often leads to nonnegligible loss of prediction accuracy \cite{bc_15,xnornet_16,bnn_16,twn_16}. Compared to BNNs, Ternary weight neural networks (TNNs) \cite{twn_16,ttq_16,twn_17} own an extra state 0 for the weights and thus enjoy a larger model capacity. TNNs benefit in the same way as BNNs do from quantization.Thanks to sparsity, a number of additions/subtractions can be further dropped from foward propagation. To store ternary numbers, we need 2-bit representation which results in $16\times$ model compression rate. TNNs strike a balance between the accuracy and memory storage. Other methods for training general low-bit CNNs have also been proposed \cite{inq_17, twn_17}.

In this paper, we study the crime forecasting at small spatial and hourly temporal scales. We adapt the ST-ResNet structure for our purposes. Compared to the traffic flow data handled by ST-ResNet, crime data is more challenging. Crime data has much less spatial temporal regularity; i.e., the number of events in adjacent time intervals and spatial cells differ hugely. Crime data are very sparse in both space and time. Crime types are also diverse \cite{RN3516}. Our contribution is four-fold. First, we select the appropriate spatial temporal scales at which crimes are predictable. We explore the suitable representation for the spatial temporal crime distribution. Second, we provide different approaches for data regularization in both spatial and temporal dimensions to further enhance the predictable signals. Third, we adapt the deep learning architecture for crime forecasting. Fourth, we study the ternarization of our ST-ResNet model.

We organize the paper as follows: In section \ref{Data}, we discuss crime data sets and preprocessing techniques. In section \ref{Algorithm}, we discuss the deep learning algorithms and network structures for crime forecasting. Forecasting results and comparisons with some other methods are presented in sections \ref{Results} and \ref{Comparison}, respectively. In section \ref{Quantization}, we explore the ternarization of the ST-ResNet to reduce the model size and speed up forecasting. In section \ref{Conclusion}, we summarize this paper's contribution and discuss future work.

\section{Data Representation} \label{Data}
\subsection{Data set description}
We consider crime forecasting in Los Angeles (LA). In our protocol, historical crime, weather and historical holiday data are the key ingredients. Since holiday records are easy to obtain, we only provide brief descriptions of the other two data sets.

\begin{figure}
\centering
\begin{tabular}{cc}
\includegraphics[width=0.49\columnwidth]{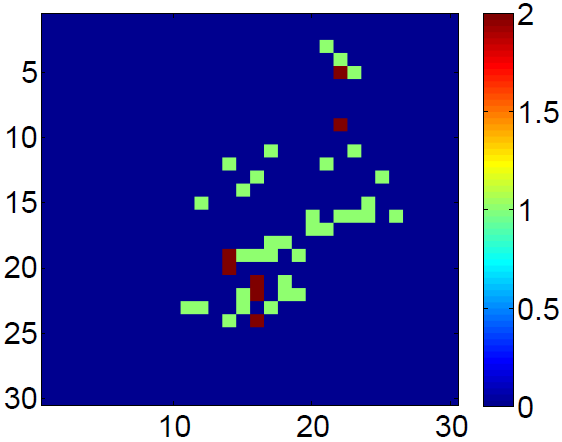}&
\includegraphics[width=0.49\columnwidth]{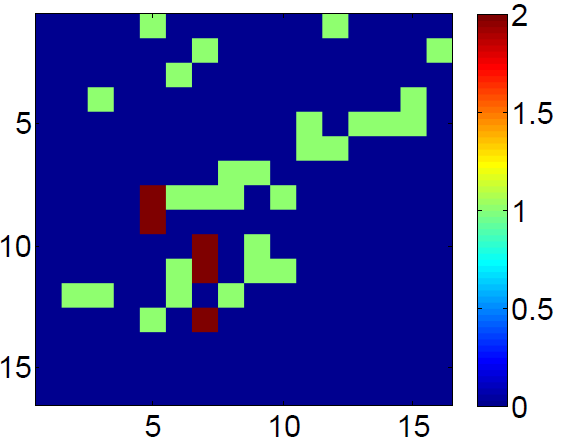}\\
(a)&(b)
\end{tabular}
\caption{Crime distribution at 1:00 p.m, Dec 20th, 2015. Chart (a) depicts crime distribution over the whole LA area; chart (b) depicts crime distribution over the selected region. The units are described in Sec. \ref{Data}.}
\label{Crime_Distribution_Demo}
\end{figure}

\paragraph{Crime Data}
For a simple yet effective demonstration of our framework, we consider all the crimes recorded in LA over the last six months of 2015 without distinguishing their types. In total there were 104,957 crime events. The crime time and location information is used in our forecasting paradigm. Each crime is associated with two times: start and end times. To avoid ambiguity, we regard the start time of each event as the associated time slot. Geographically, the latitude and longitude intervals spanned by these crimes are $[33.3427^\circ, 34.6837^\circ]$ and $[-118.8551^\circ, -117.7157^\circ]$, respectively. The spatial crime distribution is highly heterogeneous; a large portion of the area contains little or no crime. Therefore, we only consider the crimes that happened within the region $[33.6927^\circ, 34.3837^\circ]\times[-118.7051^\circ, -118.1157^\circ]$, this selected region contains more than 95 percent of the total crimes. Nevertheless, there is still spatial redundancy in this data embedding. In our study, we partition this selected region into a $16\times 16$ lattice. Each grid cell represents approximately $17.8 km^2$ land area.
Fig. \ref{Crime_Distribution_Demo} shows the crime distribution at 1:00 p.m on Dec 20th, 2015. The left panel is the crime distribution over the whole LA area. The right panel depicts the crimes in the restricted region.
All crime historical data is provided by Los Angeles Police Department (LAPD).

\paragraph{Weather Data}
We collect the weather data from the Weather Underground database available at \url{https://www.wunderground.com/} using a simple web crawler. Special attention should be paid to get extracting the data correctly as the format varies day by day. We select temperature, wind speed, and special events, including fog, rain, and thunderstorms, for our weather features. Since we study hourly crime forecasting, if more than one weather data are available, we make use of an average of the features. For the time intervals without weather data, we use a linear interpolation from neighboring intervals.

\subsection{Data Preprocessing}
\begin{figure}
\centering
\begin{tabular}{cc}
\includegraphics[width=0.49\columnwidth]{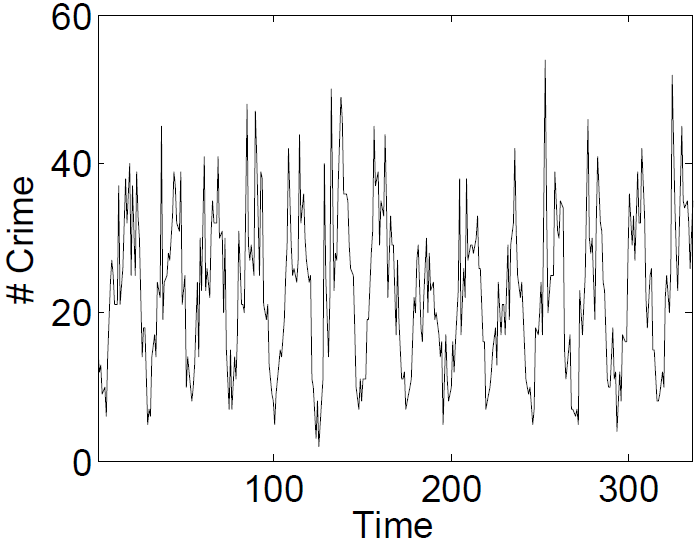}&
\includegraphics[width=0.49\columnwidth]{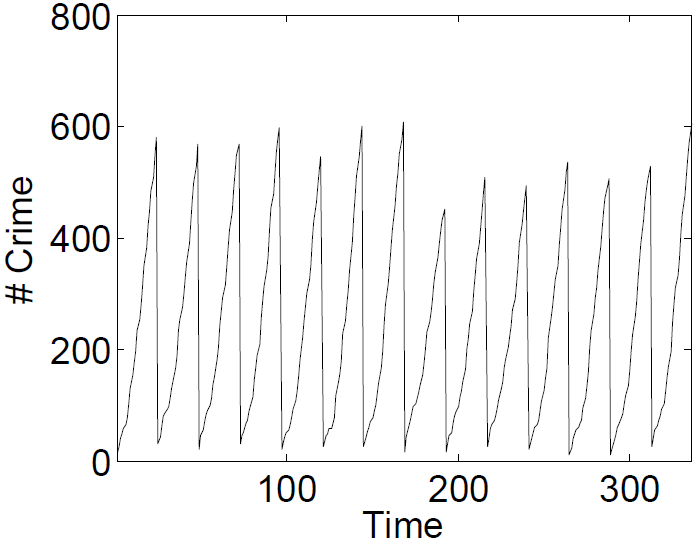}\\
(a)&(b)\\
\includegraphics[width=0.49\columnwidth]{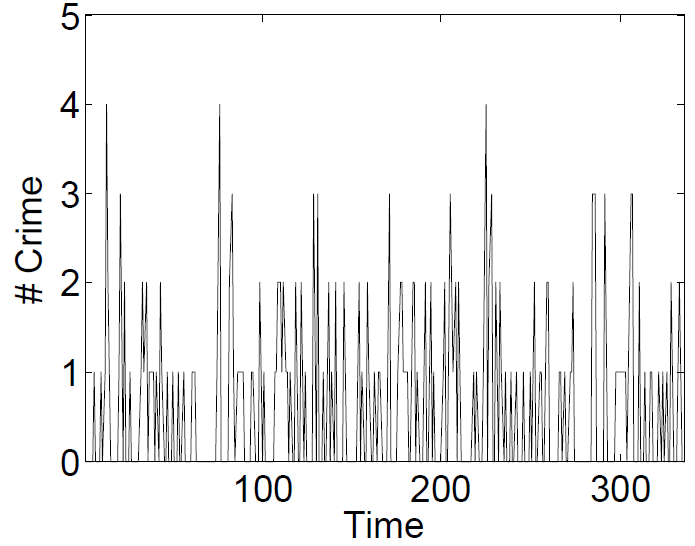}&
\includegraphics[width=0.49\columnwidth]{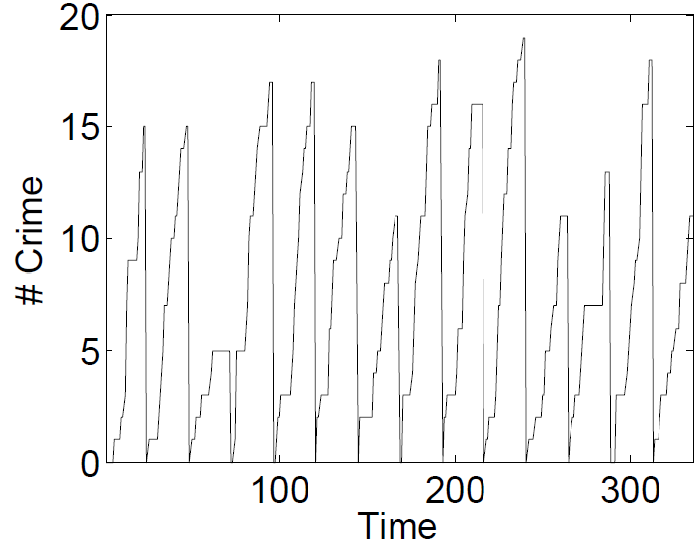}\\
(c)&(d)\\
\end{tabular}
\caption{Chart (a) depicts the hourly crime intensity of the last two weeks of 2015 over the whole LA area; chart (b) draws the cumulated crime intensity corresponding to (a).  Charts (c) and (d) plot crime density and diurnal cumulated crime intensity on the grid with longitude and latitude ranged $[33.9519^\circ, 33.9951^\circ]\times[-118.2635^\circ, -118.2262^\circ]$, respectively. Units: x-axis: time; y-axis: number of crimes.}
\label{Crime_Intensity_Demo1}
\end{figure}

Charts (a) and (c) of Fig. \ref{Crime_Intensity_Demo1} show crime intensity functions in the whole LA and a randomly selected grid over the last two weeks of the year 2015. The intensity functions show low regularity in the temporal dimension. However, the hourly crime time series indicate strongly predictable signals; obviously, the time series over the whole domain is periodic with a period of 24 hours. For selected grid cells, the periodic patterns still exists, but the time series much more irregular. Deep learning uses combinations of simple linear and nonlinear continuous functions to form a dynamical system, thus approximating the complex input signal. Since deep learning models are essentially continuous, we need to enhance the regularity of the time series data, especially for the grid-wise crime intensity functions. To address this, we map the original crime intensity function $\{X(t)\}$ to $\{Y(t)\}$ via a diurnal periodic integral mapping:
\begin{equation}
\label{cum}
Y(t)=\int_{nT}^t X(s)ds,
\end{equation}
for $t$ within the time interval $(nT, (n+1)T]$. As demonstrated in charts (b) and (d) of Fig.\ref{Crime_Intensity_Demo1}, after integration, the regularity of the original time series improves dramatically. The periodic signal is amplified.

\begin{figure}
\centering
\begin{tabular}{cc}
\includegraphics[width=0.49\columnwidth]{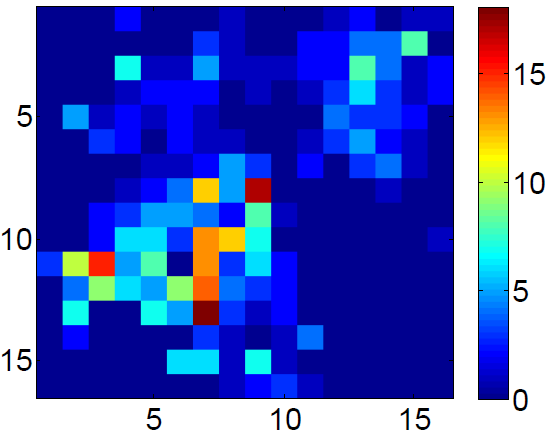}&
\includegraphics[width=0.49\columnwidth]{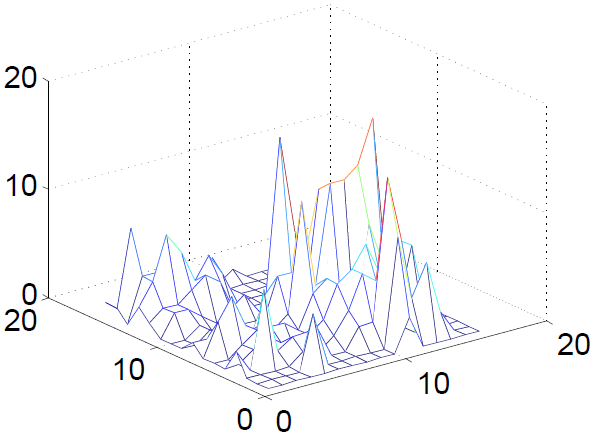}\\
(a)&(b)\\
\includegraphics[width=0.49\columnwidth]{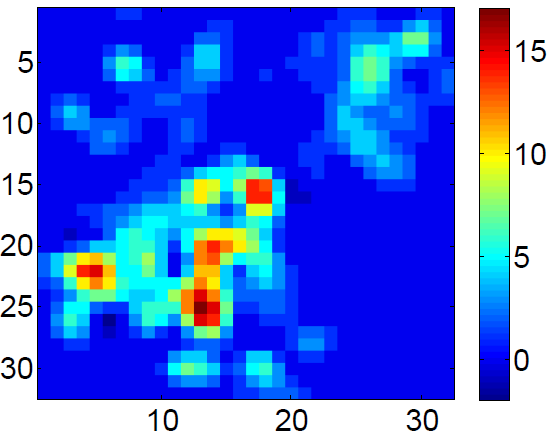}&
\includegraphics[width=0.49\columnwidth]{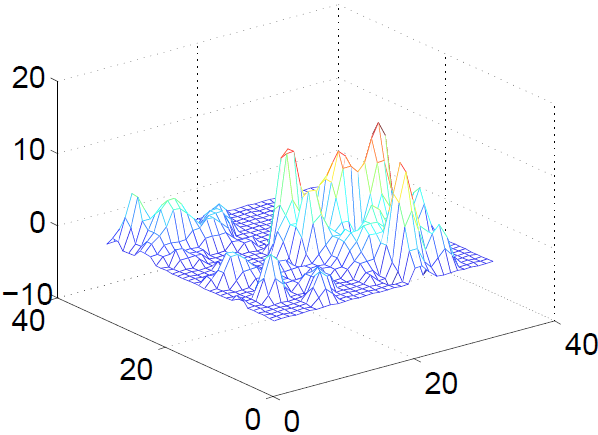}\\
(c)&(d)
\end{tabular}
\caption{Cumulated crime intensity at 11:00 p.m, Dec 31st, 2015. Chart (a) depicts crime distribution over the selected area; chart (b) provides the mesh plot of the chart (a); chart (c) depicts super resolution version of chart (a); and char (d) is mesh plot of chart (c).}
\label{Crime_cumDistribution_Demo}
\end{figure}

To resolve the lack of spatial regularity, we use a super resolution technique at each time step; e.g., bilinear and cubic spline interpolation. For computational efficiency, we resolve by a factor of 2 in each dimension of the spatial domain. In Fig. \ref{Crime_cumDistribution_Demo} we see that the bilinear spline super resolution significantly improves spatial regularity. A merit of this preprocessing is that it improves the signal without losing information associated with the crime data.

\section{Models and Algorithms} \label{Algorithm}
\subsection{Mathematical Problem Formulation}
For the sake of simplicity, in this work we do not consider the crime type forecasting problem. In our protocol, we only consider how many crimes will happen in the next time step in each grid cell. Mathematically, our paradigm can be formulated as: given the historical data $\{(X_t, E_t)\}_{t=1, 2, \cdots n}$ and future external features $\{E_{n+1}\}$, predict $X_{n+1}$, where $X_1, X_2, \cdots, X_{n+1}$ are the tensors representing the crime spatial distributions at times $t_1, t_2, \cdots, t_{n+1}$. $E_1, E_2, \cdots, E_{n+1}$ are the external features that affect the crimes, (e.g., holiday, time, weather). The entire procedures of our crime predictor can be formulated by the pseudo code described in Algorithm. \ref{CrimePredictor}. In Algorithm. \ref{CrimePredictor}, $S$ and $I$ denote spatial super-resolution and temporal diurnal integration operators, respectively.

\begin{algorithm}
\caption{Real Time Spatial Temporal Deep Learning Crime Predictor}\label{CrimePredictor}
\begin{algorithmic}[1]
\Procedure{Crime Predictor}{}
\State \textbf{Input: } Crime historical spatial distributions: $\{X_t\}_{t=1}^n$, and external features $\{E_t\}_{t=1}^{n+1}$
\State \textbf{Output: } Crime spatial distribution $X_{n+1}$ at time slot $t_{n+1}$.
\State \textbf{Step 1: } Perform spatial super resolution on the input crime distributions to get $S(X_t)$ for $t=1, 2, \cdots n$.
\State \textbf{Step 2: } Perform temporal super resolution on the spatial super resolved data to get $I(S(X_t))$ for $t=1, 2, \cdots n$.
\State \textbf{Step 3: } Train the ST-ResNet on the concatenation of the super resolved historical crime data and external features, $\{(I(S(X_t))), E_t\}_{t=1}^n$.
\State \textbf{Step 4: } Predict the spatial temporal super resolved crime distribution at time slot $t_{n+1}$ by the trained ST-ResNet and external features $E_{n+1}$.
\State \textbf{Step 5: } Temporal recovery: $I^{-1}(I(S(X_{n+1})))=S(X_{n+1})$. 
\State \textbf{Step 6: } Spatial recovery: $S^{-1}(S(X_{n+1}))=X_{n+1}$.
\State \textbf{Step 7: } Get the prediction
$$
X^p_{n+1}=\left\{
            \begin{array}{ll}
              |X_{n+1}|_+, & \hbox{if}\ \ mod(n, 24)=0 \\
              \max\{|X_{n+1}|_+, X_{n}\}, & \hbox{if}\ \ mod(n, 24)\neq 0
            \end{array}
          \right.
$$
where $|X_{n+1}|_+$ is the positive part of $X_{n+1}$, i.e.,
$$
|X_{n+1}|_+=\left\{
            \begin{array}{ll}
              0, & \hbox{if}\ \ X_{n+1}<0 \\
              X_{n+1}, & \hbox{if}\ \ X_{n+1}\geq 0
            \end{array}
          \right.
$$
\EndProcedure
\end{algorithmic}
\end{algorithm}

\subsection{ST-ResNet structure}
We test two different deep neural network structures. The first structure is adapted from \cite{Junbo:2017}. The second structure, which excludes convolution, is equivalent to  an ensemble of residual networks to learn the time series on each grid, without considering the transition of crimes between different grids. The first model is more realistic. Through convolutional layers, crime dynamics and influences among different grids can be captured. In both networks, all features are fused with the crime data via a parametric-matrix based fusion technique used in \cite{Junbo:2017}. The detailed description of the network structure can be found in \cite{Junbo:2017}. We implement our method using Keras \cite{Keras:2015} on top of Theano \cite{Theano:2016} software.
\begin{figure}[!ht]
\small
\centering
\includegraphics[width=14cm,height=12cm]{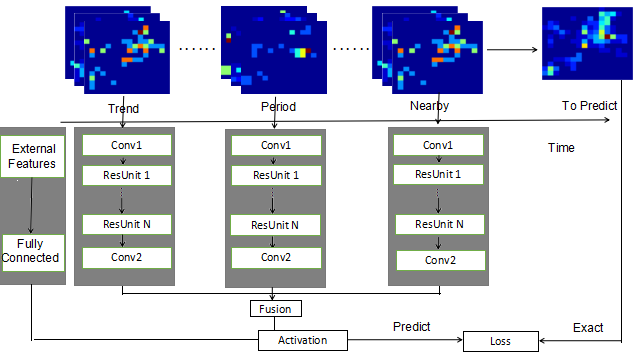}
\caption{Structure of the deep neural network model with convolution.}
\label{ConvResNet}
\end{figure}
Our models incorporate external features such as weather and holidays. Due to the periodic pattern and self-exciting property of crimes \cite{Mohler:2011JASA}, we adopt nearby, periodic, and trend features. The time spacing of these features are at hourly, daily, and weekly levels, respectively. For each category of these dependencies, we employ the three nearest previous spatial distributions of crimes. For instance, suppose we wish to predict the crime distribution at $t_{n+1}$, the past crime distributions: $X_{n}$, $X_{n-1}$, $X_{n-2}$, $X_{n-24}$, $X_{n-48}$, $X_{n-72}$, $X_{168}$, $X_{n-336}$, $X_{n-504}$ are utilized as features. We believe that longer dependencies produce better results. We let the algorithm learn the dependencies automatically in a RNN fashion.

\section{Results on Crime Forecasting} \label{Results}
We ran experiments on the last six months crime data of 2015 over LA. The last two weeks data is used to test the model. The remaining data is used for training and validating the models, where the validation ratio is 20 percent. We use 6 layers of residual units, the number selected by trial-and-error, to assemble the ST-ResNet, which is a good compromise between model complexity and accuracy. In the training period, we first run 200 epochs to train the network with a separated validation set to ensure our models do not over-fit. Subsequently, we schedule another 50 epochs on the combination of the training and validation sets to fine tune the model. All the experiments are carried out with a single Nvidia Quadro-K4000 graphics card. To speed up the training process, we make use of the deep neural network library cuDNN \cite{cuDNN:2014}. The size of the convolution filters are fixed to 3 $\times$ 3. The learning rate is chosen to be 0.0005. The ADAM optimizer is used to optimize the loss function.

We use the root mean square error (RMSE) between prediction and ground truth as our measure of accuracy of the predictions. RMSE  is defined as:

\begin{equation}
\label{RMSE:Define}
{\rm RMSE}=\sqrt{\frac{1}{N*T}\sum_{i, t}(I_{it}-I_{it}^p)^2},
\end{equation}

where $N$ is the total number of grids that we partition the restricted area into, $T$ is the number of time slots considered, $I_{it}$ and $I_{it}^p$ are the exact and predicted crime intensity in grid $i$ at time $t$, respectively. When considering the accuracy of the prediction in a single grid cell, we do not need to sum over the index $N$. Table \ref{Performance:DifferentNet} lists the RMSEs between the predictions and the ground truth cumulative intensity functions with different setups of the network and different treatments of the input data.

We consider different experimental setups to validate the importance of the signal enhancement treatments. For super resolution comparisons, we consider three cases, namely super resolution in space and time, super resolution in space only, and no super resolution. Bilinear interpolation is employed for all signal enhancements. As demonstrated in Table \ref{Performance:DifferentNet}, the best results come from using both spatial and temporal super resolutions. In general, these signal enhancement techniques improve model performance.  To test the influence of model complexity, we considered different number of filters in the convolutional layers.
We list the results for different number of neurons in Table \ref{Performance:DifferentNet} In general, performance increases with the more neurons involved. These filters capture different scales of the spatial temporal features of the training data set. Currently, the maximum number of filters (64$\times$64) is set by the capacity of our graphics card. We believe the model can give even better results with more filters, since they can capture more detailed information about the spatial temporal distribution of crime. The optimal results obtained when we use convolutional layers on the super resolved signals on both space and time, which gives RMSE 0.207 in the prediction. The performance shows that convolutional layers captures the spatial influence of crimes, as it is known that crime is self-exciting in both space and time \cite{Mohler:2011JASA}. Without convolutional layers, each grid is basically treated independently, which leads to an inefficient model.

\begin{figure}
\centering
\begin{tabular}{cc}
\includegraphics[width=0.49\columnwidth]{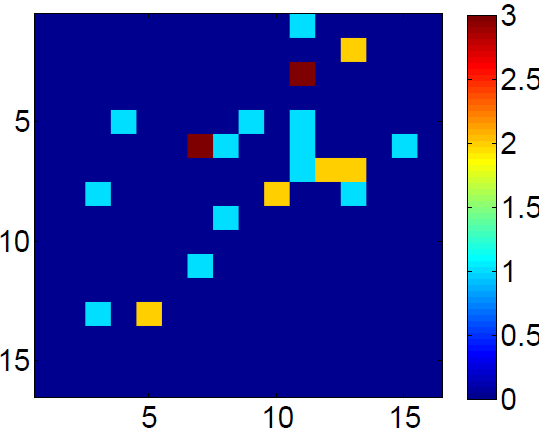}&
\includegraphics[width=0.49\columnwidth]{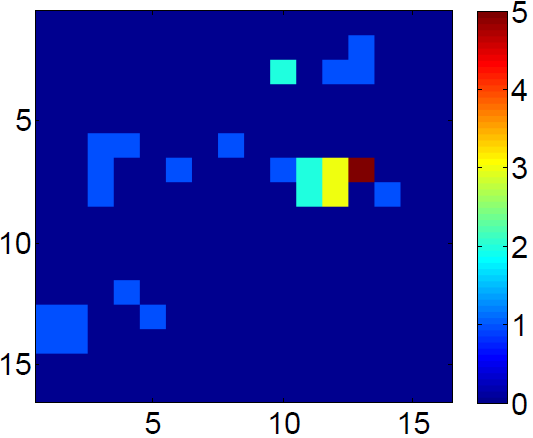}\\
(a)&(b)\\
\includegraphics[width=0.49\columnwidth]{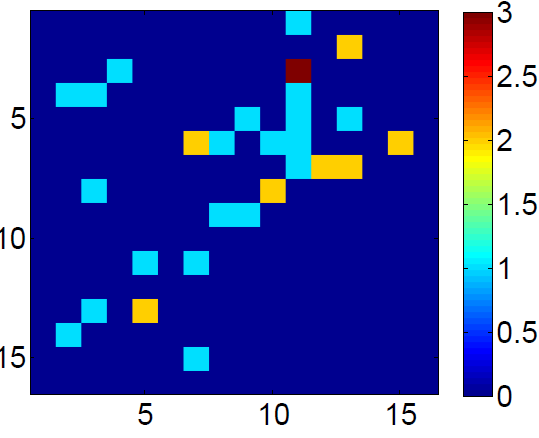}&
\includegraphics[width=0.49\columnwidth]{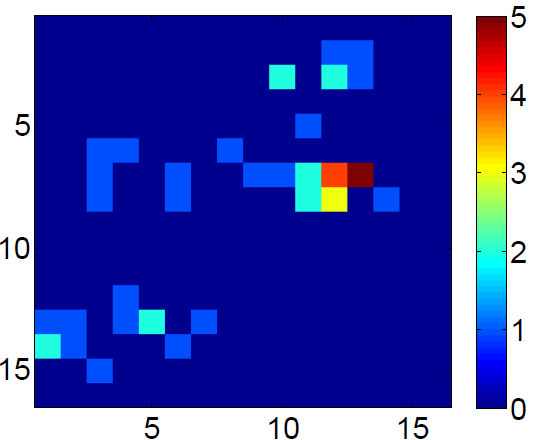}\\
(c)&(d)\\
\includegraphics[width=0.49\columnwidth]{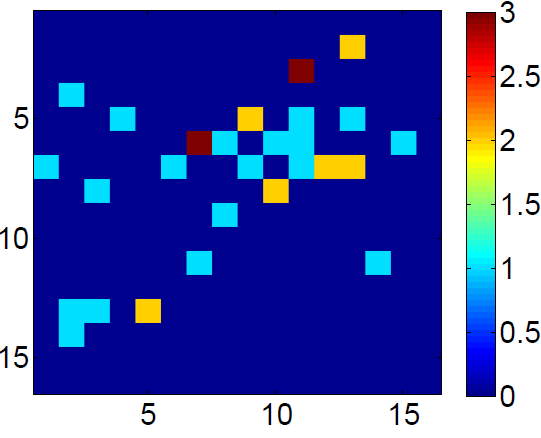}&
\includegraphics[width=0.49\columnwidth]{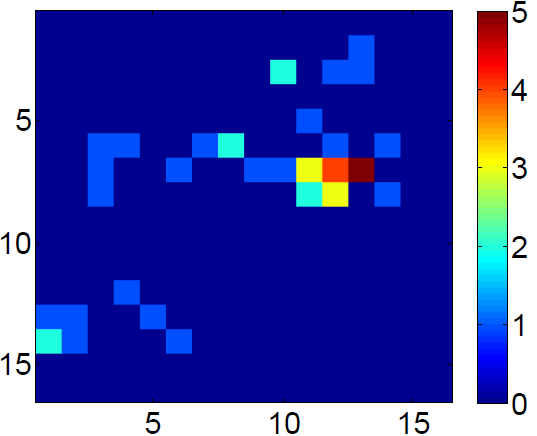}\\
(e)&(f)
\end{tabular}
\caption{Predicted vs. exact crime spatial distribution. Panels (a), (b) plot the crime spatial distribution at 1 p.m. of Dec 19, 27, 2015, respectively. Panels (c), (d) are the predicted results without convolution layers. (e), (f) are the predicted results with convolution layers.}
\label{Predicted_Crime_Distribution}
\end{figure}

In Fig. \ref{Predicted_Crime_Distribution} we show sample snapshots in time. It is easy to see all crime hot spots are captured. The ST-ResNet gives satisfactory results in the cases with or without convolutional layers.

\begin{figure}
\centering
\begin{tabular}{cc}
\includegraphics[width=0.49\columnwidth]{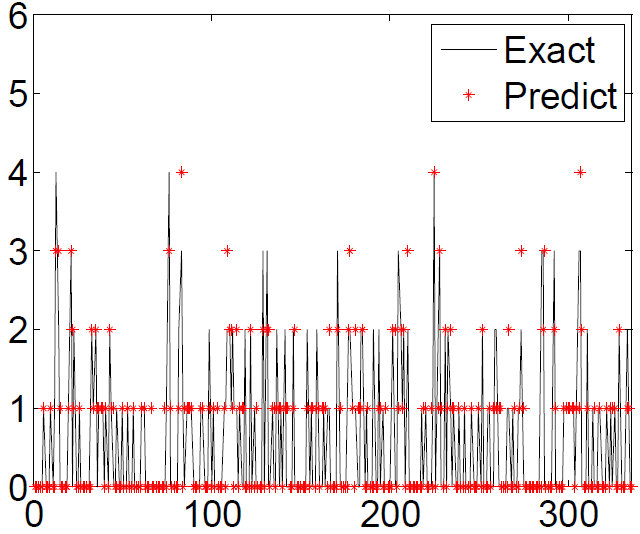}&
\includegraphics[width=0.49\columnwidth]{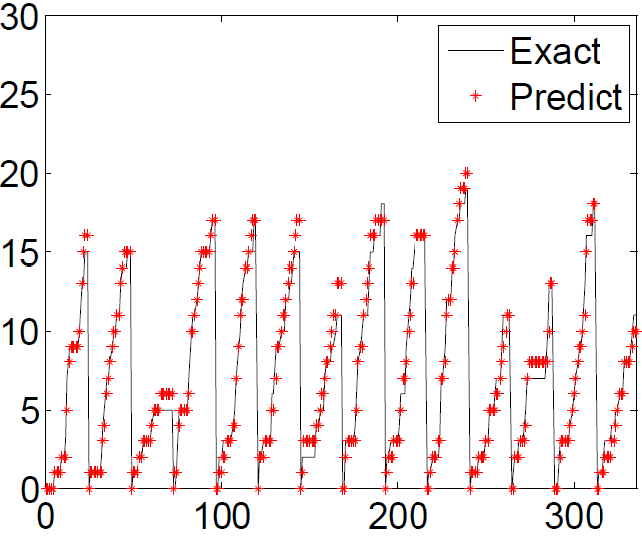}\\
(a)&(b)\\
\includegraphics[width=0.49\columnwidth]{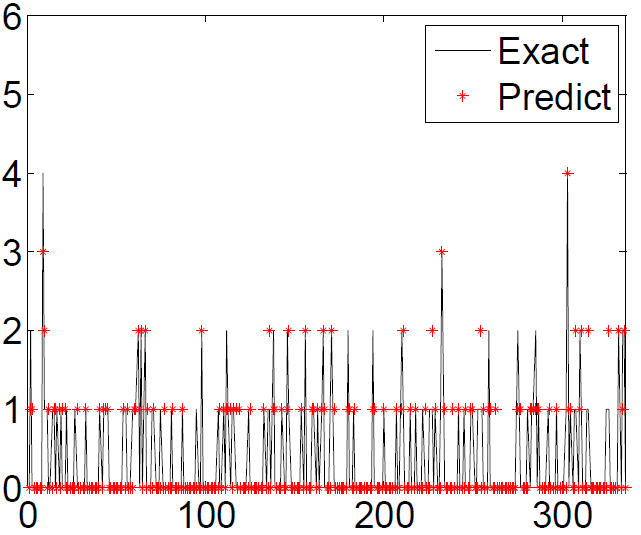}&
\includegraphics[width=0.49\columnwidth]{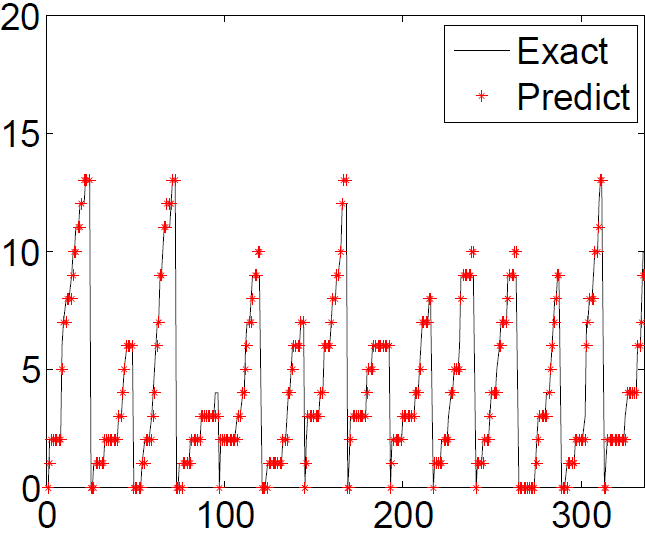}\\
(a)&(b)\\
\end{tabular}
\caption{Predicted vs. exact crime intensity in two randomly selected grid cells area over the last two weeks of 2015. Charts (a) and (b) are prediction on the crime intensity and cumulated intensity functions on the grid $[33.9519^\circ, 33.9951^\circ]\times[-118.2635^\circ, -118.2262^\circ]$, respectively. Charts (c) and (d) are the corresponding intensity and cumulated intensity prediction over the grid $[34.0382^\circ, 34.0382^\circ]\times[-118.4472^\circ, -118.4104^\circ]$. Units: x-axis: time; y-axis: number of crimes.}
\label{Predicted_Crime_TimeSeries}
\end{figure}

For a given grid, the crime intensity over a given time interval is also accurately predicted. We randomly select two grids with longitude and latitude ranges $[33.9519^\circ, 33.9951^\circ]\times[-118.2635^\circ, -118.2262^\circ]$ and $[34.0382^\circ, 34.0382^\circ]$ $\times$
$[-118.4472^\circ, -118.4104^\circ]$, respectively. As shown in Fig. \ref{Predicted_Crime_TimeSeries}, the maximum difference between the ground truth and the prediction in crime intensity is 3 crimes in absolute value. These results quantitatively confirm our predictions are accurate. The RMSE of the prediction over crime intensity functions are 0.665 and 0.551, respectively; 0.750, 0.443 over the cumulated intensity functions.
For the first grid, there are 131 hourly time slots with crimes over the last two weeks of 2015. Our predictor gives 148 candidates, the intersection with the ground truth is 106. For the second grid, there are 99 hourly time slots with crimes. The prediction gives 104 slots, 69 of them lies in the ground truth set.

One key feature of the convolutional neural network is weight sharing, i.e., for a given neuron, it shares the common filter over the whole image domain. This simplifies the neural network model and the training procedure. However, for extremely sparse spatial data, like the crime data we study, this weight sharing may lead to the filter with all weights being zeros. As shown in Table \ref{Performance:DifferentNet}, without super resolution, the network with convolutional layers offers worst forecasting (all the predictions are zero). We conclude that, for sparse spatial data, applying convolutional network to the super resolved data can give excellent forecasting. On the one hand, it solves weight sharing problem. On the other, it captures complex spatial distributions.

\renewcommand{\arraystretch}{1.5}
\begin{table}[tp]
\centering
\fontsize{8.5}{10}\selectfont
\begin{threeparttable}
\caption{Performance of ST-Resnet on the crime forecasting under different settings. Units for Training and Test Error columns: number of crimes}\label{Performance:DifferentNet}
\begin{tabular}{ccccc}
\toprule
\toprule
\multicolumn{5}{c}{With Convolution Layers}\cr
\toprule
\#of neurons    &\#of parameters   &Training Time (s)   &Training RMSE  &Test RMSE\cr
\midrule
\multicolumn{5}{c}{Spatial Temporal Super-resolution}\cr
\midrule
64 $\times$ 64  &1,350,911	       &104157.02   &0.143	         &0.207\cr
32 $\times$ 32  &350,879	       &30506.27    &0.189		     &0.231\cr
16 $\times$ 16  &99,695		       &15917.58    &0.255		     &0.298\cr
\midrule
\multicolumn{5}{c}{Spatial Super-resolution Only}\cr
\midrule
64 $\times$ 64  &1,350,911	       &49056.339	   &0.313		 &0.323\cr
32 $\times$ 32  &350,879	       &14390.37	   &0.293	     &0.361\cr
16 $\times$ 16  &99,695		       &7554.76		   &0.364		 &0.417\cr
\midrule
\multicolumn{5}{c}{No Super-resolution}\cr
\midrule
64 $\times$ 64  &1,350,911	       &11053.91		   &1.92			 &1.83\cr
32 $\times$ 32  &350,879	       &3428.59    		   &1.92			 &1.83\cr
16 $\times$ 16  &99,695		       &1910.16 		   &1.92		     &1.83\cr
\bottomrule
\toprule
\multicolumn{5}{c}{Without Convolution Layers}\cr
\toprule
\multicolumn{5}{c}{Spatial Temporal Super-resolution}\cr
\midrule
64 $\times$ 64        &165,119	    &23137.48		&0.363		  &0.379\cr
32 $\times$ 32	      &52,895		&8774.47	    &0.366		  &0.385\cr
16 $\times$ 16		  &24,431		&5801.79		&0.376		  &0.391\cr
\midrule
\multicolumn{5}{c}{Spatial Super-resolution Only}\cr
\midrule
64 $\times$ 64        &165,119	    &10905.29	    &0.379        &0.397\cr
32 $\times$ 32	      &52,895		&4162.88		&0.413		  &0.426\cr
16 $\times$ 16		  &24,431		&2743.15		&0.399		  &0.401\cr
\midrule
\multicolumn{5}{c}{No Super-resolution}\cr
\midrule
64 $\times$ 64        &165,119	    &3009.18		    &0.378    &0.397\cr
32 $\times$ 32	      &52,895		&1178.22		    &0.413	  &0.425\cr
16 $\times$ 16		  &24,431		&797.32  		    &0.396	  &0.399\cr
\bottomrule
\end{tabular}
\end{threeparttable}
\end{table}

\section{Comparison between Different Methods} \label{Comparison}
In this section, we compare our approach with several existing methods for crime forecasting. In total, we compare our deep learning approach to ARIMA  \cite{Chen:2008ARIMA}, k nearest neighbor (KNN), and historical average (HA). We brief summarize these methods in the following:
\begin{itemize}
\item HA: in this simple empirical model, at each time slot, we regard the historical average at that specific hour as the prediction. This is a parameter free model. However, the daily crime volatility cannot be captured by this model.
\item KNN: in this model, we use the average number of crimes in the closest previous time steps to forecast the number of events at the next time step. The only parameter is $k$, which represents the number of nearest previous steps involved in the prediction. The parameter $k$ can be determined by simple cross validation. Here we adopt five-fold cross validation to determine this parameter. It is found that when $k$ equal to one, KNN provides the best results. Lag forecasting is the main drawback of this model.
\item ARIMA: the general model ARIMA$(p, d, q)$ has three parameters, where $p$ is the order of autoregressive model, $d$ is the order of difference needed to make the signal to be stationary, $q$ is the order of the moving average. The parameter $d$ is determined by the ADF stationarity test, $p$ and $q$ are determined by the autocorrelation function (ACF) and partial autocorrelation function (PACF), respectively. Based on our testing, the cumulative crime intensity function itself is stationary. The optimal order for autoregression and moving average are 25 and 26, respectively. These two parameters reflect a roughly one day dependence. Due the the simplicity of training the model, we implement the ARIMA model in a rolling fashion and update the model on the fly as new data is presented. The major deficiency of this model is that it cannot included features other than the time series itself. It is also too simple to capture all the features carried by the signal. In general, ARIMA is only suitable for simple time series that carry all the predictable information.
\end{itemize}

\begin{figure}
\begin{tabular}{cc}
\includegraphics[width=0.49\columnwidth]{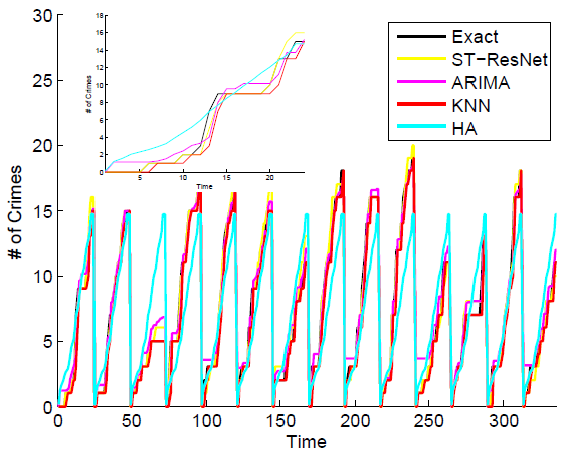}&
\includegraphics[width=0.49\columnwidth]{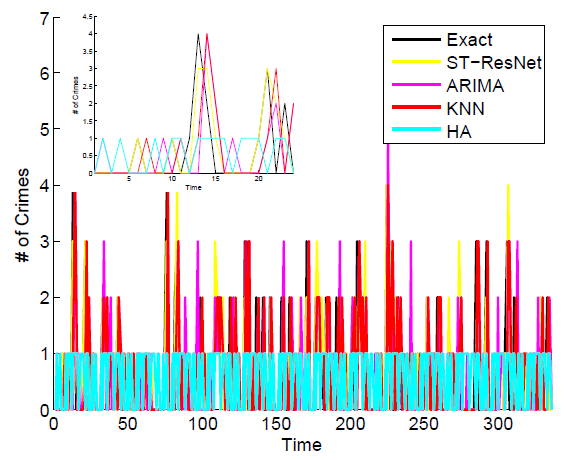}\\
(a)&(b)
\end{tabular}
\caption{Comparison of different methods' forecasting on the c.d.f and p.d.f over the last two weeks of 2015 on the region $[33.9519^\circ, 33.9951^\circ]\times[-118.2635^\circ, -118.2262^\circ]$. Charts (a) and (b) are the forecasting results of c.d.f and p.d.f, respectively. We have also provide a zoom in plot the crimes of the first day over this period.}
\label{Compare}
\end{figure}

We randomly select a grid cell with the longitude and latitude range $[33.9519^\circ, 33.9951^\circ]\times[-118.2635^\circ, -118.2262^\circ]$ for comparison. The comparison of exact and predicted cumulative densities of crime is depicted in panel (a) of Fig.\ref{Compare}. Panel (b) of Fig.\ref{Compare} is a comparison of the crime intensity functions. The crime distribution function itself is highly irregular over the time span. The regularity of the signal is enhanced by integration. The cumulative density function is periodic with some fluctuation.

The deep learning model provides the optimal prediction, followed by ARIMA, KNN, and HA (Table \ref{Performance:Comparison}).  ARIMA, KNN, HA are not on par with one another. The optimal RMSEs in cumulative density and original crime signal are 0.750 and 0.659, respectively. For ARIMA and KNN, the errors in the original signal are more than the cumulative one. Visually, ARIMA and KNN seem to provide excellent predictions. However, this is a misperception due to lagged forecasting. According to our tests, ST-ResNet shows even stronger performance relative to the other predictors when the data become more sparse.

\renewcommand{\arraystretch}{1.5}
\begin{table}[tp]
\centering
\fontsize{8.5}{10}\selectfont
\begin{threeparttable}
\caption{Performance comparison between different methods over the last two weeks of 2015 on the region $[33.9519^\circ, 33.9951^\circ]\times[-118.2635^\circ, -118.2262^\circ]$. Units for the Error columns: number of crimes}
\label{Performance:Comparison}
\begin{tabular}{ccc}
\toprule
\toprule
Method    &Error in cumulated crime density   &Error in crime density  \cr
\midrule
ST-Resnet           &0.750	         &0.659\cr
fully-ternary ST-Resnet &0.612       &0.705\cr
ARIMA(25, 0, 26)    &1.115		     &1.213\cr
KNN(1)              &1.168		     &1.236\cr
HA                  &2.618           &1.568\cr
\bottomrule
\end{tabular}
\end{threeparttable}
\end{table}

\section{Ternarization of ST-ResNet}\label{Quantization}
In this section, we consider the ternarization of ST-ResNet. Suppose there are in total $l$ fully-connected and convolutional layers with the respective weight filters $W_i$, $i = 1,\dots, l$. For a fully-connected layer, $W_i$ is a matrix, and for a convolutional layer, it is a high dimensional tensor. Without loss of generality, let us view $W_i$ as a vector of dimension $n_i$. Then the vector $W_i\in \R^{n_i}$ is ternary-valued and takes the form $$W_i = \alpha_i T_i,$$
where $T_i\in \{-1,0,1\}^{n_i}$ has the same size as $W_i$, and $\alpha_i>0$ is a \emph{shared} scaling factor. Training TNNs calls for solving the following constrained minimization problem
\begin{equation}\label{model:tnn}
\min_{W, b} \; f(W,b) \quad \mbox{subject to} \quad W_i\in\T_i, \; i = 1,\dots,l,
\end{equation}
where
$f$ denotes the overall energy function determined by the network architecture, $W=\{W_1,\dots, W_l\}$ the weight parameters, $b$ the other trainable parameters, and
$$
\T_i:= \{W_i\in\R^{n_i}: \exists \; \alpha_i>0 \mbox{ and } T_i\in\{-1,0,1\}^{n_i} \mbox{ such that } W_i = \alpha_i T_i\}
$$ the set of ternary weights for the $i$-th layer.

The key step for solving Eq.(\ref{model:tnn}) lies in the ternarization of some given floating-point vector $\W_i$. To this end, we seek to minimize the Euclidean distance between $\W_i$ and $W_i$:
 $$
 \proj_{\T_i}(\W_i) :=\arg\min_{W_i} \|W_i - \W_i\|^2 \quad \mbox{subject to} \quad W_i\in\T_i.
 $$
The solution $\proj_{\T_i}(\W_i)$ to the above problem is simply the projection of $\W_i$ onto the set $\T_i$. For now let us ignore the subscript $i$ for notational simplicity. In an alternative form, the above problem can be formulated as
\begin{equation}\label{model:twn}
(\alpha^*,T^*) = \arg\min_{\alpha, T} \; \|\alpha T - \W\|^2 \quad \mbox{subject to} \quad \alpha>0, \; T\in\{-1,0,1\}^{n}.
\end{equation}
After obtaining $(\alpha^*,T^*)$, the ternarization of $\W$ is then given by $\proj_{\T}(\W) = \alpha^* T^*$. The solution to (\ref{model:twn}) was first approximated by Li et al\cite{twn_16} under unrealistic statistical assumptions on the components of $\W$, albeit with satisfactory empirical performance. The exact expression for $\proj_{\T}$ was later derived by Yin et al\cite{twn_17}. We summarize the result in the theorem below.

\begin{theorem}\label{thm:prox}
Suppose $\W_{[k]}$ keeps the $k$ largest entries in magnitude of $\W$ and zeros out the others. Then the solution to problem (\ref{model:twn}) is given by
$$
\alpha^* = \frac{\|\W_{[k^*]}\|_1}{k^*}, \quad T^* = \sgn(\W_{[k^*]}),
$$
where $k^* = \arg\max_k \; \frac{\|\W_{[k]}\|_1^2}{k}$ is the sparsity of the optimal ternary weight vector.
\end{theorem}
For readers' convenience, we provide a proof here.

\begin{proof}
Suppose the sparsity of $T$ is $k$. Since $T\in\{-1,0,1\}^n$, then
$$
\|T\|^2 = k \quad \mbox{and} \quad |\langle T,\W \rangle| \leq \|\W_{[k]}\|_1,
$$
and thus
\begin{align}\label{obj}
\|\alpha T-\W\|^2& = k\alpha^2 -2\langle T,\W \rangle\alpha  + \|\W\|^2  = k\left(\alpha - \frac{\langle T, \W\rangle}{k}\right)^2 -\frac{\langle T, \W\rangle^2}{k} + \|\W\|^2 \notag \\
& \geq  -\frac{\langle T, \W\rangle^2}{k} + \|\W\|^2 \geq -\frac{\|\W_{[k]}\|_1^2}{k} + \|\W\|^2.
\end{align}
 Since $\|\W\|^2$ is a constant, the optimal sparsity $k^*$ maximizes the term $\frac{\|\W_{[k]}\|_1^2}{k}$ in (\ref{obj}), i.e.,
\begin{equation*}\label{min_k}
k^* = \arg\max_k \; \frac{\|\W_{[k]}\|_1^2}{k}.
\end{equation*}
To achieve the lower-bound in (\ref{obj}), we must have
\begin{equation*}\label{min_T}
T^* = \sgn(\W_{[k^*]}), \; \alpha^* = \frac{\langle T^*, \W\rangle}{k^*} = \frac{\|\W_{[k^*]}\|_1}{k^*}.
\end{equation*}
\end{proof}

According to Theorem \ref{thm:prox}, the ternarization of $\W$ can be performed in a manner of direct enumeration. This involves sorting the magnitudes of elements of $\W$ and computing accumulative sum of the sorted sequence, which require computational complexity of $O(n\log(n))$.
Our training of ternary ST-ResNet is carried out by a projected SGD-like algorithm \cite{xnornet_16}. We keep updating the floating-point weights using the minibatch (sub)gradient of $f$ evaluated at ternary weights. This is different from the standard projected SGD in which the ternary weights are updated in the descent step. The mean convergence of this pseudo projected SGD has been proved under smoothness and convexity assumptions on $f$ \cite{quantization_17}. In fact, it has demonstrated much stronger empirical performance than the standard version in training quantized neural networks  \cite{quantization_17}. In addition, we adopt popular techniques in deep learning such as $\ell_2$ regularization, batch normalization \cite{bnorm_15} and ADAM \cite{adam_15}  to improve training efficiency. Our method for training ternary ST-ResNet is summarized in Algorithm \ref{alg}.

We coded and tested the optimizer on Lasagne/Theano\cite{Theano:2016,lasagne} platform in Python on a machine with Nvidia GeForce GTX Titan X GPU. We trained fully-ternary ST-ResNet with spatial and temporal super-resolution preprocessing, where all the weight filters are ternary. The training RMSE and testing RMSE with 64$\times$64 neurons are 0.234 and 0.242, respectively. As shown in Table \ref{Performance:TernaryNet}, compared to the full-precision model, there is only a small accuracy loss.

\begin{algorithm}
\caption{Training one epoch of ternary weight ST-ResNet.}\label{alg}
\begin{algorithmic}
    \State \textbf{Input:} Energy function $f(W,b)$, the number of mini-batches $d$, adaptive learning rate $\eta_{t}$, $t = 1, \dots, d$,  parameters $W^0$, $b^0$ and $\W^0$ output from the last epoch.
    \State \textbf{Output:} floating-point weights $\W^d$, ternary weights $W^d$ and other parameters $b^d$.
    \State \textbf{Initialization:} Randomly shuffle the samples in training dataset.
    \For {t = 1, $\dots$, d}
    \State $f(W^{t-1}, b^{t-1})=$EvalGrad$(f,W^{t-1},b^{t-1})$ \quad $//$ Evaluate the (sub)gradient of $f$ at $(W^{t-1},b^{t-1})$ using the $t$-th mini-batch by back propagation.
    \State $\W^{t}$ = ADAM($\W^{t-1}, \nabla_{t-1} f(W^{t-1}, b^{t-1})$, $\eta_{t-1}$)  \quad $//$ Update the floating-point weights $\W$ using ADAM.
    \For {i = 1, $\dots$, l}
    \State $W_i^{t} = \proj_{\T_i}(\W^{t}_i)$ \quad $//$ Ternarize the weights $\W_i$ in the $i$-th layer by solving (\ref{model:twn}) .
    \EndFor
    \State $f(W^{t}, b^{t-1})=$EvalGrad$(f,W^t,b^{t-1})$ \quad $//$ Evaluate the (sub)gradient of $f$ at $(W^t,b^{t-1})$ using the $t$-th mini-batch by back propagation.
    \State $b^{t}$ = ADAM($b^{t-1}, \nabla_{t-1} f(W^t, b^{t-1})$, $\eta_{t-1}$)  \quad $//$ Update other  trainable parameters $b$ using ADAM.
    \EndFor
\end{algorithmic}
\end{algorithm}

\renewcommand{\arraystretch}{1.5}
\begin{table}[tp]
\centering
\fontsize{8.5}{10}\selectfont
\begin{threeparttable}
\caption{Performance comparison between ST-ResNet and its ternarization. Units for Training and Test Error columns: number of crimes}
\label{Performance:TernaryNet}
\begin{tabular}{ccccc}
\toprule
\toprule
Model   &\#of neurons    &\#of parameters      &Training Error  &Test Error\cr
\midrule
ST-ResNet                   &64 $\times$ 64  &1,350,911	&0.143	         &0.207\cr
Fully-ternary ST-Resnet     &64 $\times$ 64  &1,365,031	        &0.234	     & 0.242 \cr
\bottomrule
\end{tabular}
\end{threeparttable}
\end{table}

\section{Concluding Remarks} \label{Conclusion}
In this paper, we present a real-time spatial temporal predictor for end-to-end crime intensity prediction. The key idea of our predictor can be summarized as follows:
\begin{itemize}
\item We chose appropriate spatial temporal scales at which crime historical time series carry sufficient predictable signals. For a given time step, we map the number of events into an image, each pixel value represents the number of crime in a grid at a specific time.

\item We developed effective spatial temporal signal enhancement techniques to boost the crime forecasting accuracy. These techniques also solve the deficiency of the CNNs for sparse data dues to the weight sharing. More specifically, in the temporal dimension, we compute the diurnal cumulative crime per grid spatial region. In the spatial dimension, we use bilinear interpolation super resolution.

\item We adapted the ST-ResNet for crime prediction.
\end{itemize}

Our methods provide crime forecasting for each grid cell at hourly temporal scale. The predictions are extremely accurate, which provides reliable guidance for crime control. Our model can be categorized as a deep learning regression method, which provides a better description of crime forecasting than the classification type of methods, since crime prediction is not just a simple yes-or-no problem.

Nevertheless, there are many aspects to improve. On the one hand, the ad hoc grid partitioning of the spatial domain ignores demographic and geographic information. Furthermore, embedding the irregular geometry of the city into a rectangular domain leads to a huge amount of redundant computation. On the other hand, in the ST-ResNet framework, the historical dependencies need to be set explicitly and longer explicit dependencies cause the network to be extremely complex and difficult to train. Adaptive dependence is hard to incorporate into the ST-ResNet framework.

There are a few lines of research worth exploring in the future. First, a better graphical representation of spatial temporal data representation will benefit both the capturing of information from  historical data and efficient computation. Second, instead of explicitly pointing out the dependencies, an alternative is to use RNN to learn the dependencies automatically. Third, forecasting crime types is feasible in our framework, although challenging, since the data will be much more sparse compared to the present representation.

\section*{Acknowledgments}
This work was supported by ONR grants N00014-16-1-2119, N000-14-16-1-2157,  ARO MURI grant W911NF-11-1-0332, NSF grants DMS-1118971, DMS-1417674, DMS-1522383, and IIS-1632935.
The authors thank the Los Angeles Police Department for providing the crime data for this paper.

\end{document}